\documentclass[11pt,a4paper]{article}
\usepackage{times,latexsym}
\usepackage[T1]{fontenc}
\usepackage{xcolor}
\usepackage{multirow}
\usepackage{graphicx}
\usepackage{tabularx}
\usepackage{subcaption}
\usepackage{caption}
\usepackage{algorithm}
\usepackage{algorithmic}
\usepackage{booktabs}
\usepackage{url}
\usepackage{microtype}
\usepackage{xspace} 
\usepackage{eacl2021}

\aclfinalcopy
\def\aclpaperid{1036}

\usepackage{amsmath,amsfonts,bm}

\def\eqref#1{equation~\ref{#1}}

\def\1{\bm{1}}

\def\funf{{\pmb{f}}}

\def\va{{\bm{a}}}

\def\vg{{\bm{g}}}
\def\vh{{\bm{h}}}

\def\vs{{\bm{s}}}

\def\vx{{\bm{x}}}
\def\vy{{\bm{y}}}

\def\mX{{\bm{X}}}
\def\mY{{\bm{Y}}}

\DeclareMathAlphabet{\mathsfit}{\encodingdefault}{\sfdefault}{m}{sl}
\SetMathAlphabet{\mathsfit}{bold}{\encodingdefault}{\sfdefault}{bx}{n}

\newcommand{\softmax}{\mathrm{softmax}}

\newcommand{\npi}{\textsc{NPI}\xspace}
\newcommand{\simt}{\textsc{SiMT}\xspace}
\newcommand{\mre}{\textsc{Read}\xspace}
\newcommand{\mwr}{\textsc{Write}\xspace}
\newcommand{\nmt}{\textsc{NMT}\xspace}

\newcolumntype{L}{>{\arraybackslash}X}

\title{Learning Coupled Policies for Simultaneous Machine Translation \\ using Imitation Learning}

\author{
Philip Arthur$^\dagger$ \and Trevor Cohn$^\ddagger$ \and Gholamreza Haffari$^\dagger$ \\
$\dagger$ Department of Data Science and Artificial Intelligence\\
Monash University, Australia \\
\texttt{\{philip.arthur, gholamreza.haffari\}@monash.edu} \\
$\ddagger$ School of Computing and Information Systems \\
University of Melbourne, Australia\\
\texttt{tcohn@unimelb.edu.au} 
}

\date{\today}

\begin{document}
\maketitle
\begin{abstract}

We present a novel approach to efficiently learn a simultaneous translation model with coupled programmer-interpreter policies. 
First, we present an algorithmic oracle to produce oracle \mre/\mwr actions for  training bilingual sentence-pairs using the notion of word alignments. 
This oracle actions are designed to capture  enough information from the partial input before writing the output. 
Next, we perform a coupled scheduled sampling to effectively mitigate the exposure bias when learning both policies jointly with imitation learning. 
Experiments on six language-pairs show our method outperforms strong baselines in terms of translation  quality while keeping the translation delay low.

\end{abstract}

\section{Introduction}

Simultaneous machine translation (\simt) is a  setting where the translator needs to incrementally generate the translation while the source utterance is being received. 
This is a challenging translation scenario as  the \simt \ model needs to trade off  delaying translation output and the quality of the generated translation. 

Recent research on \simt relies on a strategy to decide when to read a word from the input or write a word to the output \cite{sp16,gu-etal-2017-learning}.  
This is based on a sequential decision making formulation of \simt, where the decision making about the next \mre/\mwr action is made by an \emph{agent}, interacting with the neural machine translation (NMT) \emph{environment}.
Current approaches are sub-optimal as they either fix the agent's policy  to focus learning the NMT model \cite{ma-etal-2019-stacl,dalvi-etal-2018-incremental} or learn  adaptive agent policies while the NMT model is fixed \cite{gu-etal-2017-learning,alinejad-etal-2018-prediction}.
We argue that the interpreter should also learn to generate correct translation from incomplete input information. 
This is challenging as we need to optimize both programmer's and interpreter's policies to  balance the tradeoff between quality and delay in the  reward. 

Previous research has considered the use of imitation learning (IL) to train the agent's policy \cite{DBLP:journals/corr/abs-1909-01559,DBLP:conf/acl/ZhengZMH19}, which is generally superior to reinforcement Learning (RL) in terms of the stability and sample complexity.
However, the bottleneck of IL in \simt is the \emph{un}availability of the \emph{oracle} sequence of actions. Designing \emph{algorithmic} oracles to compute sequence of \mre/\mwr actions with low translation latency and high translation quality is under-explored. 

We present an IL approach to efficiently learn effective coupled programmer-interpreter policies in \simt, based on the following contributions. 
First, we  present  a simple, fast, and effective algorithmic  oracle  to  produce  oracle actions from the training bilingual sentence-pairs based on statistical word alignments \cite{Brown1994}.
Next, we design a framework that uses scheduled sampling on both programmer and interpreter. 
This is different from the typical IL scenarios, where there is only one policy to learn.
As the two policies \emph{collaborate}, their learning needs to be  robust not only to their own incorrect predictions, but also to  incorrect predictions of the other policy to mitigate this coupled  \emph{exposure bias}.

Experiments on six language pairs (translating to English from Arabic, Czech, German, Romanian, Hungarian, and Bulgarian) show the policies trained using our approach 
compares favorably with strong policies from the previous work.
We attribute the effectiveness of the learned coupled policies to (i) the scheduled sampling, which handles the coupled exposure bias, resulting in up to 5-8 BLEU score improvements, and (ii) the quality of oracle actions generated by our algorithmic oracle, which balances translation quality and delay.

\section{\npi \  Approach to \simt}

We describe generation in our neural programmer-interpreter (\npi) approach to  simultaneous machine translation (\simt) in Algorithm \ref{alg:npi}.
At each time step $t$, the \emph{programmer} needs to decide whether to \mre the next source word  or to \mwr the next target word in the translation. 
The \emph{interpreter} then immediately \emph{executes} the action generated by the programmer. 
Both programmer and interpreter are modeled using Markov Decision Process (MDP) where prediction at particular timestep depends on the history of previous predictions. The indices $i_t$  and $j_t$  are the number of \mre and \mwr actions in the program up to time step $t$.

\begin{algorithm}[t]
\caption{Generation in \npi-\simt}
\begin{algorithmic}[1]
\STATE $i,j \leftarrow 0$

\WHILE{a stopping condition is not met} 
\STATE $t \leftarrow i+j$
\STATE $\vs_{t+1} \leftarrow \funf_{\textrm{prog}}(\vs_{t}, [a_{t},\vg_j, \vh_i])$
\STATE $P_{\textrm{prog}} \leftarrow \softmax(\text{dense}_{\textrm{prog}}(\vs_{t+1}))$
\STATE $a_{t+1} \sim  P_{\textrm{prog}}$
\IF{$a_{t+1} = \textrm{\mre}$}
\STATE $i \leftarrow i + 1$
\STATE $\vh_i \leftarrow \funf_{\textrm{enc}}(\vh_{i-1},x_i)$ 
\ELSE 
\STATE $j \leftarrow j + 1$
\STATE $\vg_j \leftarrow \funf_{\textrm{intp}}(\vg_{j-1},y_{j-1},\vh_{\leq i})$
\STATE $P_{\textrm{intp}} \leftarrow \softmax(\text{dense}_{\textrm{intp}}(\vg_{j}))$ 
\STATE $y_j \sim P_{\textrm{intp}}$ 
\ENDIF
\ENDWHILE
\end{algorithmic}
\label{alg:npi}
\end{algorithm}

\paragraph{The Programmer} needs to \emph{sequentially} decide about the next \emph{action}, given the previous actions $\va_{<t}$ and the prefix of the source utterance read so far $\vx_{\leq i_t}$ as well as the prefix of the target translation generated so far $\vy_{\leq j_t}$. That is, our programmer is modeled as $P_{\textrm{prog}}(a|\va_{<t},\vx_{\leq i_t}, \vy_{\leq j_t})$.


\paragraph{The Interpreter} needs to execute the action generated by the programmer. 
At time step $t$, if the generated action $a_t$ is \mre, we reveal the next input token. Otherwise, if \mwr, then we generate the next target word according to $P_{\textrm{intp}}(y|\va_{\leq t},\vx_{\leq i_t}, \vy_{\leq j_t})$.\footnote{The counter $i$ and $j$ are also incremented according to the respective actions at time $t$.}



\paragraph{The Probabilistic Model.}

The probability of simultaneously generating the translation $\vy$ and the sequence of actions $\va$ for a source utterance $\vx$ is,
\vspace{-4mm}
\begin{eqnarray*}
P_{\textrm{\simt}}(\vy,\va|\vx) = 
 \prod_{t=1}^{|\vx|+|\vy|} P_{\textrm{prog}}(a|\va_{<t},\vx_{\leq i_t}, \vy_{\leq j_t}) \\ 
 \times \prod_{t: a_t=\textrm{\mwr}} P_{\textrm{intp}}(y|\va_{\leq t},\vx_{\leq i_t}, \vy_{\leq j_t}).
\end{eqnarray*}


\paragraph{Training the Model.}
In \simt, we are interested in not only producing a high quality translation, but also reducing the delay between the times of receiving the source words and generating their translations. 
Training of the model based on this hybrid training objective can be done by  reinforcement learning (RL) or imitation learning (IL). 
%
%
The RL approach has been attempted by \cite{sp16,gu-etal-2017-learning,alinejad-etal-2018-prediction} for training the programmer; 
however, it is unstable due to sparsity of the reward function and these works also assumed a fixed interpreter. 
%
%
We thus take the IL approach for a sample efficient, effective, and stable learning of policies in   \npi-\simt.

\section{Deep Coupled Imitation Learning}

Our goal is to learn a pair of policies for the programmer and interpreter using IL.
\S\ref{subsec:couple} describes the method of learning of both policies where their learning inter-dependency needs to be taken into account. 
\S \ref{subsec:oracle} describes our novel \emph{oracle} program actions for each sentence pair in the training set, i.e., the program $\va$ which has been responsible for generating the translation $\vy$ for a 
source utterance $\vx$ with as low delay as possible. Our overall training algorithm is depicted in Algorithm 2. 
$\hat{\mX} = [\hat{\vx}_1, ..., \hat{\vx}_{|\vx|}]$ is the encoding of the input sequence and $\hat{\mY} = [\hat{\vy}_1, ..., \hat{\vy}_{|\vy|}] $ is a list of interpreter hidden states for all predictions. During training, these values are calculated before calculating the loss of the programmer.

\subsection{Learning Robust Coupled Policies}
\label{subsec:couple}

Assuming we have the oracle actions, we  can learn the policies for both the programmer and interpreter using behavioural cloning in IL \cite{ijcai2019-882}. 
That is, the model parameters are learned by  maximising the likelihood of the oracle actions for both the programmer and interpreter, 
\begin{eqnarray*}
\theta_{\textrm{prog}}^*,\theta_{\textrm{intp}}^* :=  &\arg\max_{\theta_{\textrm{prog}},\theta_{\textrm{intp}}}  \sum_{(\vx,\vy,\va)} \\ \sum_{t=1}^{|\vx|+|\vy|}  & \log P_{\textrm{prog}}(a|\va_{<t},\vx_{\leq i_t}, \vy_{\leq j_t};\theta_{\textrm{prog}}) \\
+ \sum_{t: a_t=\textrm{\mwr}} 
& \log P_{\textrm{intp}}(y|\vx_{\leq i_t}, \vy_{\leq j_t}; \theta_{\textrm{intp}}).
\end{eqnarray*}
This is akin to have the expectation, in the original training objective of \npi, under a point-mass distribution over the oracle actions.

IL with  behavioural cloning does not lead to robust policies for unseen examples in the test time due to exposure bias \cite{NIPS2015_5956}.  
%
That is, the agent is only exposed to situations resulting from the \emph{correct} actions in the training time, leading to its inability to mitigate from propagation of errors faced due to incorrect actions in the test time. 
%
Scheduled sampling \cite{NIPS2015_5956,pmlr-v15-ross11a} addresses this issue by exposing the agent to incorrect decisions in training time through perturbation of  the oracle decisions, which we extend to learning policy \emph{pairs}. 
Crucially, the programmer-interpreter policies need to be robust to incorrect decisions encountered not only in their own trajectories, but also to  one anothers' trajectories.

%


\paragraph{Learning the Programmer.}
To train our programmer on a training example $(\vx,\vy,\va)$ with scheduled sampling, we first create  the  perturbation $(\va',\vy')$ of the ground truth program and interpreter decisions.
The perturbed program $\va'$ and translation $\vy'$ are only used as the input to the recurrent architectures of the programmer and interpreter's decoder. 
They are created by replacing some of the ground truth element by randomly selecting an action from the predictive distribution of each model. 
We then maximise the following training objective,
\begin{eqnarray}
\label{eqn:trainProg}
\theta_{\textrm{prog}}^* :=  &\arg\max_{\theta_{\textrm{prog}}}  \sum_{(\vx,\vy',\va, \va', \va'')} \sum_{t=1}^{|\vx|+|\vy'|} \nonumber \\ 
& \log P_{\textrm{prog}}(a|\va'_{<t},\vx_{\leq i_t}, \vy'_{\leq j_t};\theta_{\textrm{prog}}). 
\end{eqnarray}
Based on the generative process described in Algorithm \ref{alg:npi}, the programmer conditions the generation of actions in each time step on the current states of the NMT's encoder and decoder. 
Hence, while training the programmer, the valid \mre/\mwr actions need to be communicated to the interpreter and be executed in order to provide NMT's encoder/decoder states to the programmer to condition upon. 
Crucially, the communicated program needs to be valid.

\begin{figure}[t!]
    \centering
    \includegraphics[scale=0.9]{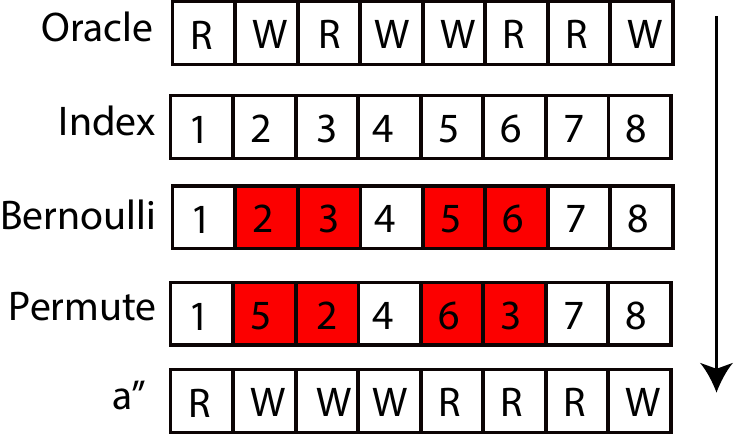}
    \caption{Creating valid perturbation from oracle program. We use a combination of Bernoulli sample and permutation function.}
    \label{fig:oracle_program}
\end{figure}

\paragraph{Valid Program} A ground truth program $\va$ is a valid sequence of \mre/\mwr actions if $|\mre \in \va| =|\vx|$ and $|\mwr \in \va| = |\vy|$. This valid program ensures the \npi model to safely consume a pair of parallel sentence. We generate a valid perturbation $\va''$ by only permuting the \mre/\mwr actions of the program $\va$ (Figure \ref{fig:oracle_program}). We further extend the definition of a valid program with respect to the domain knowledge of translation so that: (i) no \mwr at the beginning, and (ii) no \mre at the end of the program.

\begin{algorithm}[t]
\caption{Training \npi-\simt}
\label{alg-train}
\begin{algorithmic}[1]
\REQUIRE $\mathcal{D}$: Sentence pairs with oracle actions, $\beta_1, \beta_2, \beta_3$ : scheduled sampling probabilities for $\vy', \va', \va''$.

\WHILE{a stopping condition is not met} 
\STATE randomly pick $(\vx,\vy,\va) \in \mathcal{D}$
\STATE $\vy' \leftarrow \textrm{perturbSeq}(\vy,\beta_1,\theta_{\textrm{intp}})$
\STATE $\va' \leftarrow \textrm{perturbSeq}(\va,\beta_2,\theta_{\textrm{prog}})$
\STATE $\va'' \leftarrow \textrm{perturbProgValid}(\va, \beta_3)$

\STATE $\hat{\vy}, \hat{\mX}, \hat{\mY} \leftarrow \textrm{forward\_intp}(\theta_{\textrm{intp}}, \vx, \vy', \va'')$
\STATE $\hat{\va} \leftarrow \textrm{forward\_prog}(\theta_{\textrm{prog}}, \va', \hat{\mX}, \hat{\mY})$

\STATE $\theta_{\textrm{intp}} \leftarrow \theta_{\textrm{intp}} - \alpha_1 \nabla \delta(\hat{\vy}, \vy)$ 
\STATE $\theta_{\textrm{prog}} \leftarrow \theta_{\textrm{prog}} - \alpha_2 \nabla \delta(\hat{\va}, \va)$ 

\ENDWHILE
\end{algorithmic}
\end{algorithm}

\paragraph{Learning the Interpreter.} The interpreter needs to be robust to the incorrect actions in the previously generated words in the translations as well as the \mre/\mwr actions generated by the programmer. This is done by communicating $\va''$ to the intepreter during training. Thus, the training objective for the interpreter is, 
\begin{eqnarray}
\label{eqn:trainIntp}
\theta_{\textrm{intp}}^* :=  &\arg\max_{\theta_{\textrm{intp}}}  \sum_{(\vx,\vy,\vy',\va'')} 
\sum_{t: a''_t=\textrm{\mwr}} \nonumber \\
& \log P_{\textrm{intp}}(y|\vx_{\leq i_t}, \vy'_{\leq j_t}; \theta_{\textrm{intp}}).
\end{eqnarray}

\subsection{Oracle Program Actions}
\label{subsec:oracle}

Our proposed oracle should measure the appropriate amount of inputs needed for translating a particular target word $\vy_{j_t}$.
This is done by determining the \emph{key} word or phrases $\delta_j$ which contain important information of $\vy_{j_t}$, and therefore guiding the programmer to read until $\delta_j$ before writing $\vy_{j_t}$.
Algorithm \ref{alg:oracle} outlines our oracle generation procedure. 
No \mre operation is emitted if the $\delta_j \leq \delta_{\mre}$ or $\{i \in a_{i,j}\} = \emptyset$.

$\delta_j$ can be heuristically determined by using word alignment \cite{Brown1994, koehn-etal-2003-statistical}, which captures strong relationship between tokens. 
In the case of many to one alignment of source to target, we choose the furthest source word.
In the case of no alignment, nothing is done as it means the target word can be induced merely from the decoder without needing to read additional inputs. 
This oracle can generally generate a valid program. Caution is needed to ensure that no \mwr at the beginning and no \mre at the end of the generated oracle. 
This can be done by aligning the first words of the parallel sentence. Similarly we also need to align the last words of the parallel sentence.
\footnote{Our oracle algorithm's code is released in \url{https://github.com/Monash-NLP-ML-Group/arthur-eacl2021}.}

\begin{algorithm}[t]
\caption{Oracle Generation}
\begin{algorithmic}[1]
\REQUIRE $\va$: Symmetrized alignment of $\vx$ and $\vy$ in forms of $\va_{i,j}$, which means that $\vx_i$ is aligned to $\vy_j$. Index starts with 0. 
\STATE $\delta_{\mre}:= -1$ 
\FOR{$j \in \textbf{range}(0, |\vy|)$}
    \STATE $\delta_j$ $\leftarrow \max [\{i \in a_{i,j}\}]$
    \FOR {($\delta_j - \delta_{\mre}$) times}
        \STATE \textbf{emit}  \mre
    \ENDFOR
    \STATE $\delta_{\mre} \leftarrow \max(\delta_j, \delta_{\mre})$
    \STATE \textbf{emit} \mwr
\ENDFOR
\end{algorithmic}
\label{alg:oracle}
\end{algorithm}
\section{Experiments} 

Our experiments aim to measure the effectiveness of our proposed method versus a strong wait-$k$ baseline, over a range of languages of varying difficulty, syntactic complexity, and lexical complexity.

\newcommand*\rot[1]{\rotatebox{90}{#1}}
\setlength\tabcolsep{2pt} 
\begin{table*}[ht!]

\begin{subtable}{\textwidth}

\centering

\resizebox{\columnwidth}{!}{
\begin{tabular}{l||cccc|cccc|cccc|cccc|cccc|cccc}
& \multicolumn{4}{c}{DE$\rightarrow$EN} 
& \multicolumn{4}{c}{CS$\rightarrow$EN} 
& \multicolumn{4}{c}{AR$\rightarrow$EN}
& \multicolumn{4}{c}{HR$\rightarrow$EN} 
& \multicolumn{4}{c}{BG$\rightarrow$EN} 
& \multicolumn{4}{c}{RO$\rightarrow$EN}
\\

& BLEU & DAL & AL & AP & BLEU & DAL & AL & AP & BLEU & DAL & AL & AP 
& BLEU & DAL & AL & AP & BLEU & DAL & AL & AP & BLEU & DAL & AL & AP \\

\toprule 
wait-$k$ \\ 

\ \ \ $k=1$  
& \underline{14.40} & \underline{3.74} & \underline{2.22} & \underline{0.61} & \underline{11.87} & \underline{4.05} & \underline{2.60} & \underline{0.63} & \underline{16.08} & \underline{4.04} & \underline{2.88} & \underline{0.64} 
& \underline{24.78} & \underline{3.72} & \underline{2.15} & \underline{0.58} & \underline{22.02} & \underline{3.58} & \underline{1.58} & \underline{0.56} & \underline{22.76} & \underline{3.51} & \underline{1.43} & \underline{0.55}
\\

\ \ \ $k=2$ 
& \underline{18.67} & \underline{4.01} & \underline{2.93} & \underline{0.64} & \underline{15.90} & \underline{4.21} & \underline{3.20} & \underline{0.66} & \underline{19.46} & \underline{4.35} & \underline{3.44} & \underline{0.68} 
& \underline{28.50} & \underline{3.99} & \underline{2.80} & \underline{0.60} & \underline{25.37} & \underline{3.99} & \underline{2.27} & \underline{0.59} & \underline{28.31} & \underline{3.82} & \underline{2.13} & \underline{0.58}
\\

\ \ \ $k=3$ 
& \underline{21.13} & \underline{4.59} & \underline{3.71} & \underline{0.68} & \underline{17.27} & \underline{4.82} & \underline{3.92} & \underline{0.70} & 22.11 & 4.91 & 4.21 & 0.72 
& \underline{32.04} & \underline{4.55} & \underline{3.52} & \underline{0.63} & \underline{26.79} & \underline{4.90} & \underline{3.35} & \underline{0.62} & \underline{32.13} & \underline{4.31} & \underline{2.88} & \underline{0.61}
\\

\ \ \ $k=4$ 
& 24.21 & 5.03 & 4.32 & 0.71 & \underline{18.54} & \underline{5.49} & \underline{4.72} & \underline{0.74} & 22.60 & 5.67 & 5.04 & 0.75 
& 35.16 & 5.20 & 4.36 & 0.66 & \underline{31.42} & \underline{5.37} & \underline{4.10} & \underline{0.65} & \underline{34.74} & \underline{5.12} & \underline{3.83} & \underline{0.64}
\\

\ \ \ $k=5$ 
& 25.63 & 5.81 & 5.19 & 0.75 & 19.77 & 6.23 & 5.59 & 0.77 & 23.22 & 6.50 & 5.90 & 0.79
& 37.42 & 5.91 & 5.12 & 0.69 & 34.83 & 5.85 & 4.73 & 0.67 & 38.70 & 5.63 & 4.37 & 0.66 
\\

\ \ \ $k=6$ 
& 26.19 & 6.68 & 6.11 & 0.78 & 20.19 & 7.00 & 6.41 & 0.80 & 23.09 & 7.47 & 6.90 & 0.82 
& 38.41 & 6.90 & 6.14 & 0.72 & 37.26 & 6.64 & 5.56 & 0.70 & 39.90 & 6.63 & 5.59 & 0.69 
\\

\ \ \ $k=7$ 
& 26.89 & 7.51 & 7.01 & 0.81 & 20.12 & 7.95 & 7.33 & 0.83 & 23.51 & 8.20 & 7.67 & 0.84 
& 39.77 & 7.69 & 6.97 & 0.74 & 38.42 & 7.53 & 6.54 & 0.72 & 40.35 & 7.54 & 6.46 & 0.72 
\\

\ \ \ $k=\infty$ 
& 28.05 & 21.67 & 21.67 & 1.00 & 20.85 & 20.91 & 20.91 & 1.00 & 23.54 & 19.98 & 19.98 & 1.00
& 41.86 & 27.56 & 27.56 & 1.00 & 41.98 & 29.06 & 29.06 & 1.00 & 46.22 & 29.88 & 29.88 & 1.00 
\\
\midrule

\npi-\simt 
& 17.53 & 5.05 & 1.96 & 0.57 & 13.58 & 3.23 & 1.16 & 0.55 & 15.78 & 3.65 & 1.34 & 0.57
& 25.72 & 4.30 & 1.82 & 0.55 & 25.42 & 5.75 & 2.61 & 0.57 & 24.60 & 5.81 & 2.46 & 0.57 
\\

\ \ \ $+ \va'$ 
& 20.89 & 3.98 & 1.65 & 0.56 & 16.17 & 3.02 & 1.22 & 0.55 & 16.12 & 2.88 & 1.14 & 0.56
& 30.66 & 4.05 & 1.83 & 0.55 & 30.73 & 4.25 & 1.96 & 0.55 & 30.47 & 5.07 & 2.25 & 0.56 
\\

\ \ \ $+ \va''$ 
& 20.83 & 5.14 & 2.04 & 0.58 & 16.96 & 4.07 & 1.56 & 0.57 & 17.18 & 3.44 & 0.96 & 0.56
& 33.45 & 5.31 & 2.35 & 0.58 & 32.97 & 6.80 & 3.25 & 0.60 & 34.05 & 7.45 & 3.43 & 0.61 
\\

\ \ \ $+ \va', \va''$ 
& \underline{\textbf{22.37}} & \underline{\textbf{4.11}} & \underline{\textbf{1.83}} & \underline{\textbf{0.57}} & \underline{\textbf{18.69}} & \underline{\textbf{3.32}} & \underline{\textbf{1.43}} & \underline{\textbf{0.56}} & 19.33 & 3.20 & 1.31 & 0.57 
& \underline{\textbf{33.69}} & \underline{\textbf{4.23}} & \underline{\textbf{2.00}} & \underline{\textbf{0.56}} & \underline{\textbf{33.78}} & \underline{\textbf{4.69}} & \underline{\textbf{2.18}} & \underline{\textbf{0.56}} & \underline{\textbf{35.97}} & \underline{\textbf{5.32}} & \underline{\textbf{2.55}} & \underline{\textbf{0.58}}
\\

\ \ \ $+ \va', \va'', \vy'$ 
& \underline{\textbf{22.38}} & \underline{\textbf{4.04}} & \underline{\textbf{1.80}} & \underline{\textbf{0.57}} & \underline{\textbf{18.97}} & \underline{\textbf{3.24}} & \underline{\textbf{1.32}} & \underline{\textbf{0.56}} & \underline{\textbf{20.47}} & \underline{\textbf{2.89}} & \underline{\textbf{1.15}} & \underline{\textbf{0.55}}
& \underline{\textbf{35.38}} & \underline{\textbf{3.96}} & \underline{\textbf{1.84}} & \underline{\textbf{0.56}} & \underline{\textbf{34.67}} & \underline{\textbf{4.72}} & \underline{\textbf{2.26}} & \underline{\textbf{0.56}} & \underline{\textbf{37.92}} & \underline{\textbf{4.98}} & \underline{\textbf{2.38}} & \underline{\textbf{0.57}}
\\

\midrule
Oracle-at-test 
& 30.53 & 3.66 & 1.72 & 0.57 & 23.17 & 3.17 & 1.50 & 0.56 & 25.63 & 3.10 & 1.49 & 0.57 
& 44.98 & 3.84 & 1.84 & 0.55 & 46.68 & 3.98 & 1.92 & 0.56 & 50.31 & 4.32 & 2.10 & 0.56
\\
\bottomrule

\end{tabular}
}

\end{subtable}
\caption{Full results on IWSLT and SETIMES datasets. Boldface indicates better translation quality versus wait-$k$ are about the same delay (relevant systems indicated using underline within same column). Oracle-at-test is the system where the correct program is given during testing, serving as an  upper bound on translation quality at a given delay.}
\label{tab:main}
\end{table*}

\subsection{Settings} 

\paragraph{Datasets.} 
Our main experiment will be performed in higher quality corpus which is designed for spoken dialogue and carefully edited dataset. Additionally, we perform a single large scale experiment using crawled corpus such as WMT to show that our method also scales to a large dataset.

We evaluate our proposed method on 6 language pairs, in all cases translating into English, with the source languages chosen to cover a wide range of language families and syntax. 
We use German (DE), Czech (CS) and Arabic (AR) from the IWSLT 2016 translation dataset \cite{cettoloEtAl:EAMT2012}.  We use the provided training and development sets as-is, and concatenate all provided test sets to create our test set. 
We also evaluate Hungarian (HR), Bulgarian (BG), and Romanian (RO) from the SETIMES corpus \cite{tyers2010south}. As this corpus is not partitioned, we use the majority of the data for training, holding out 2000 random sentence pairs for development and another 2000 sentence pairs for testing. 
Together these languages are representative of Germanic, West Slavic, Arabic, Uralic, East Slavic, and Italic language families, respectively.

We use sentencepiece \cite{kudo2018sentencepiece} to build and tokenize our training data with 16k vocabulary size. 
Then we generate our oracle program actions based on the segmented tokens. 
We use fast\_align \cite{dyer-etal-2013-simple} to generate symmetrized alignments between tokens.
Unless otherwise specified, we use the default settings of the mentioned toolkit.


\paragraph{Evaluation.}
We evaluate the \simt systems based on its translation quality and delay. Translation quality can be measured by case sensitive BLEU \cite{papineni-etal-2002-bleu}.\footnote{Calculated using sacrebleu \cite{post2018call}. BLEU+case.mixed+numrefs.1+smooth.exp+tok.13a+ version.1.4.4 is our \texttt{sacreblue}'s signature}  
We adopt three delay measurements by previous studies. First, average proportion (AP) \cite{gu-etal-2017-learning} is a fraction of read source words per emited target words. Second, average lagging (AL) \cite{ma-etal-2019-stacl} is an average number of lagged source words until all inputs are read. Finally the differentiable-AL (DAL) \cite{arivazhagan-etal-2019-monotonic} is a refinement of AL which also accumulates the cost of writing output tokens after inputs are fully read.

%


\paragraph{Baseline.} 
We compare against the wait-$k$ baseline \cite{ma-etal-2019-stacl} where the programmer's policy begins with $k$ numbers of \mre, and is followed by switching \mwr and \mre, until the source sentence is exhausted or end of sentence (EOS) symbol is written. If the source sentence is exhausted, the programmer will only emit \mwr actions. This baseline was shown to be superior compared to the reinforcement learning approach \cite{DBLP:journals/corr/abs-1909-01559}, and $k$ can be tuned for the desired delay. \newcite{arivazhagan-etal-2019-monotonic}'s approach is superior than the wait-$k$ baseline. However there is currently no open source code available and their end-to-end approach is not using an oracle policy. As our goal is not to beat the state-of-the-art, we leave this comparison as a future work.

\paragraph{\npi-\simt.} 
Both the programmer and interpreter are modelled using a unidirectional recurrent neural network (RNN) with a long short term memory cell (LSTM). In particular, we follow the architecture of \newcite{luong-etal-2015-effective} with the multilayer perceptron attention of \newcite{DBLP:journals/corr/BahdanauCB14}.
Both the programmer and interpreter employ 20\% dropout to the network output and 10\% dropout to the embedding vector, and use a single layered LSTM with 512 hidden units. For the large scale experiment, we are using the transformer architecture  (described in \S\ref{subsec:large_scale}).

\paragraph{Training.} 
We use \texttt{Adam} optimizer \cite{DBLP:journals/corr/KingmaB14} to train this framework. We track the learning rate of programmer and interpreter separately. We start with 0.001 learning rate, and start halving it whenever perplexity increase on development set. We use a fixed perturbation probability of 5\%, 15\%, and 15\% for $\vy'$, $\va'$, and $\va''$ respectively. Early stopping is executed at the fourth learning rate decay. 

\paragraph{Testing.}
We use a beam search algorithm with a beam size of 5 and length normalization algorithm that divides hypothesis score by its length during search \cite{murray-chiang-2018-correcting}.

\begin{figure*}[ht!]
\centering
\hspace{-4.3em}\includegraphics[width=1.0\linewidth]{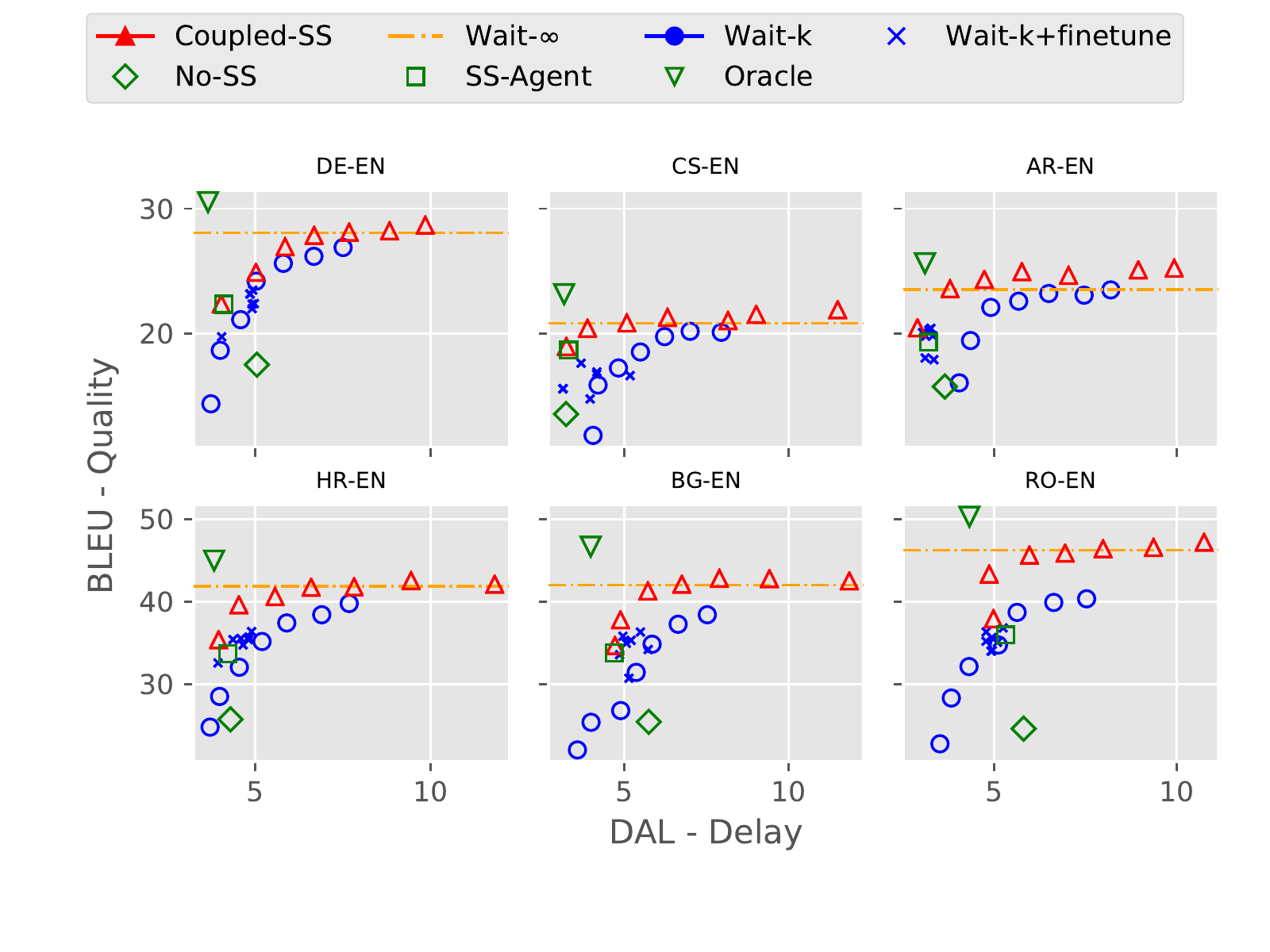}
\vspace{-1.3em}
\caption{
BLEU score versus delay using added delay and finetuning. 
Our proposed method coupled scheduled-sampling (Coupled-SS) method performs better than the wait-$k$ baseline in all settings. 
The leftmost $\triangle$ is the result of our proposed method without added delay, while $\triangle$s to the right include delay (see description in text). 
The $\Diamond$ and $\square$ are an ablation study considering not doing SS,  and doing SS only on the agent, respectively. 
The $\times$ report finetuning various pretrained wait-$k$ models with oracle+SS. 
}
\label{fig:main}
\end{figure*}

\subsection{Empirical Results}

\paragraph{Scheduled Sampling.} 

Our first experiment tests the effect of scheduled sampling (SS) in learning coupled policies in our \npi-\simt method.
For this purpose, we train four versions of our models where apply SS to both the programmer and the interpreter, only programmer, only interpreter, or neither. %
Table \ref{tab:main} shows the results, comparing against policies trained using the baseline wait-$k$ method where $k \in \{ 1,2,\ldots,7,\infty\}$. 
The \npi-\simt system that is trained using our proposed oracle (\npi\simt) is able to learn from the low delay oracle as their natural delays (DAL, AL, AP) are generally as low as the delay of the oracle during training. 
However, it is clearly difficult to perfectly predict the oracle during test-time, and these programmer prediction errors resulted in mistakes in interpreter decisions. 

Next, we consider the effect of perturbation and scheduled sampling on the proposed method.
Table \ref{tab:main} shows that applying valid perturbation ($a''$) is more important than doing normal scheduled sampling ($a'$).
This perturbation is directly correlated with the training of the interpreter, as such the noisy program make the interpreter resilient to the exposure bias.
Applying both scheduled sampling further increased our proposed method accuracy.
The schedule sampling on programmer ($+a', a''$) during training ameliorate this; as it increased up to 10 points of BLEU score in case of Romanian and Hungarian, 8 points in case of Bulgarian and, 5 points for German, Czech and Arabic.
Additionally applying scheduled sampling on the interpreter ($+y'$) further improves translation accuracy while slightly decreasing the delay, both effects being consistent across all language pairs.

\paragraph{Oracle Policy  vs Wait-$k$ Policy.} Figure \ref{fig:main} compares the policies trained by our algorithmic oracle vs those trained using the wait-$k$ policy starting from $k=1$.
In each of these six plots, the policy trained using the oracle actions corresponds to the leftmost triangle point on the figure.
Observe that the policy trained using the oracle actions compares favorably with those trained using the wait-$k$ method in terms of translation quality (higher is better) and  translation delay (lower is better). 

Next we investigate the effect of increasing the delay of the oracle policy in a controlled manner onto the translation quality of the trained systems.  
As such, we increase the delay of the oracle policy by moving the last \mre action in the oracle program to the beginning of the program, and thus increasing the delay of the oracle artificially.  
For additional delay, we repeat this process. 
We expect that the delayed oracle programs lead to trained policies with better translation quality at increased delay.
The triangles in Figure \ref{fig:main} correspond to policies trained using the versions of the oracle program, where we added delays $\{0-5\}$.
Observe that policies trained with the delayed versions of the oracle program consistently outperform the wait-$k$ policies, across all languages.

The quality of the oracle is shown as the green triangle in Figure \ref{fig:main}.
This system is provided the oracle program at test time, unlike the other systems that allow errors to propagate from the interpreter into subsequent decisions of the programmer.
Note both the low delay of the oracle, and also the fact that the BLEU score outperforms offline translation (wait-$\infty$).
This seemingly surprising finding can be explained by the oracle providing key information to the interpreter in the form of word and phrase segmentation of the inputs.

Achieving oracle level quality with a learned programmer is particularly difficult, which can be attributed to exposure bias.
However, our coupled SS method manages to bridge much of the gap between the learned program and the oracle program.

\begin{table*}[t!]
\begin{subtable}{\textwidth}
\footnotesize
\centering

\resizebox{\columnwidth}{!}{
\begin{tabular}{|l|l|l|l|l|l|l|l|l|l|l|l|l|l|l|l|l|l|l|l|l|l|l|l|}

\hline 

\_Aber \_wenn & \_wir & \textcolor{red}{\_die} & \textcolor{red}{\_Zusammensetzung} & \textcolor{red}{\_des} & \textcolor{red}{\_Erd} & \textcolor{red}{boden} & \textcolor{red}{s} & \textcolor{red}{\textbf{\_nicht}} & \textbf{\textcolor{red}{\_\"andern}} & ,& \_werden & \_wir & \_das & \_nie & \_tun & . & </s> \\

\_ & B & ut & \_if & \_we & \textcolor{red}{\textbf{\_look}} & \textcolor{red}{\_at} & \textcolor{red}{\_the} & \textcolor{red}{\_composition} &  \textcolor{red}{\_of} & \textcolor{red}{\_} & \textcolor{red}{E} & \textcolor{red}{ar} & \textcolor{red}{th} & \textcolor{red}{'} &\textcolor{red}{s} & \textcolor{red}{\_ground} & , \_we \_never \_will .\\

\hline
\end{tabular}
}

\end{subtable}
\begin{subtable}{\textwidth}

\centering

\resizebox{\columnwidth}{!}{
\begin{tabular}{|l|l|l|l|l|l|l|l|l|l|l|l|l|l|l|l|l|l|l|l|l|l|l|l|}
\hline

\_Aber & \_wenn & \_wir & \textcolor{blue}{\_die \_Zusammensetzung \_des \_Erd boden s \textbf{\_nicht} \textbf{\_\"andern}} , & \_werden \_wir & \_das \_nie & \_tun . & </s> \\

\_ & B & ut & \textcolor{blue}{\_if \_we \textbf{\_don ' t \_change} \_the \_composition \_of} & \textcolor{blue}{\_the} & \textcolor{blue}{\_soil} , & \_we  & \_will \_never \_do \_that .\\ 

\hline
\end{tabular}
}

\end{subtable}
\begin{subtable}{\textwidth}

\centering

\resizebox{\columnwidth}{!}{
\begin{tabular}{|l|l|l|l|l|l|l|l|l|l|l|l|l|l|l|l|l|l|l|l|l|l|l|l|}
\hline

\_Aber & \_wenn & \_wir & \textcolor{blue}{\_die \_Zusammensetzung \_des \_Erd boden s \textbf{\_nicht}} & \textcolor{blue}{\textbf{\_\"andern}} , & \_werden \_wir & \_das \_nie & \_tun . </s> \\

\_But & \_if & \_we & \textcolor{blue}{\textbf{\_don ' t}} & \textcolor{blue}{\textbf{\_change} \_the \_composition \_of \_the} & \textcolor{blue}{\_soil} , \_we \_will & \_never & \_do \_this .\\ 

\hline
\end{tabular}
}

\end{subtable}
\caption{The comparison of our wait-$k$, our coupled-SS (Co-SS) and the oracle trajectory. A column shows a sequence of consecutive \mre and \mwr. Here our proposed method is able to imitate the oracle well, by patiently waiting for sufficient input to produce a good translation. Red texts indicate place of translation error.}
\label{fig:deen_comp}
\end{table*}

\paragraph{Finetuning Wait-$k$.} Next we consider warm-starting training using the wait-$k$ model, and finetuning with the oracle program and SS training method. 
This method of training is cheaper when a wait-$k$ system is available, as training  converges in few iterations.
In this setting, we allow retraining of the interpreter, as fixing the interpreter  yielded poor BLEU scores.
The $\times$ points in Figure \ref{fig:main} show the results of this experiment, where each point was warm-started with a different wait-$k$ system. 
As it can be seen, all of these runs achieve similar results, but with inferior delay and quality to our proposed method (leftmost $\triangle$).
One explanation to this result is that the interpreter already converged to the wait-$k$ policy and retraining it results in an inferior model compared to training it jointly from scratch.%
\footnote{
Note that we also try to finetune the wait-$k$ system without scheduled sampling but it yields far worse performance than the one with scheduled sampling. 
This finding is similar with the main experiment.
}

\subsection {Qualitative Analysis}
\label{subsec:qualitative_analysis}

\newcite{gu-etal-2017-learning} address the difficulty of translating sentences in subject-object-verb order when translating from German to English.
We show a typical example in Table \ref{fig:deen_comp} where the wait-$k$ systems are forced to make difficult decisions with insufficient evidence, in this case of predicting negation of a verb which appears latter in the input.
We compare systems with $AL\approx2$ and $DAL\approx4$ which is achieved by the wait-2, our coupled-SS system, and the oracle system.
Consider first the wait-2 system. 
From the state shown at bottom right corner of the smaller red box, the system next generates a poor choice of verb (``look''), which was done without access to the verb in the German input.
Instead the rightmost  context word was ``Zusammensetzung'' (``composition''), which gives little information about the verb.
One way around this problem is to use a specialized classifier which predicts the final verb \cite{grissom-ii-etal-2014-dont}.
However, this is often onerous or impossible.
In this example, the model must also predict the negation ``nicht'' which appears immediately before the final verb.
In a real interpretation scenario, the only way to ensure  we output a correct translation is to wait for the matrix verb and negation token. 

In this example the oracle trajectory breaks down the input sentence into coherent chunks, and this leads to excellent translation of each segment, and with low delay.
This is because the word alignment oracle includes crossing alignment inside a phrase, thus producing sequence of \mre and \mwr that do not break phrase translations.
We posit that this oracle provides the minimal  context needed for \simt to translate on the fly, with sufficient context to generate each output token.

Here our proposed system closely imitates the oracle trajectory.
Our proposed method is more conservative in waiting for the input, waiting until the final verb to make precise prediction.
This can be explained by the uncertainty over breaking the phrases by the programmer, and thus is incurs additional delay for the sake of better translation quality.
Such behaviour that is observed in the output of human interpreters, who will often wait when they are unsure what the main speaker is talking about.

\subsection{Oracle Behaviour towards Alignment}

Section \ref{subsec:qualitative_analysis} has partly shown our oracle behavior in translating from a final-verb language into English. Here we discuss the oracle's action when translating into a final verb language (English-Dutch).
In English, the past participle is usually found right after the auxiliary verb.
In this case, our oracle actions are conservative when waiting for inputs on the target side.

\begin{figure}[t]
    \centering
    \fbox{\includegraphics[scale=0.8]{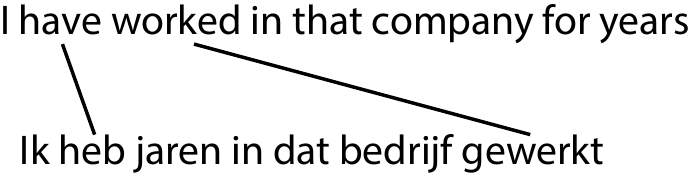}}
    \caption{Translation example from English to Dutch. In this case ``have worked'' produce a non crossing alignment with  ``heb ... gewerkt'' in Dutch.}
    \label{fig:en_nl}
\end{figure}

The example is shown in Figure \ref{fig:en_nl} in which ``have worked'' does not produce a crossing alignment. 
First, the generated oracle will \mre ``have'' and \mwr ``heb''. 
Then it will examine the next word, ``jaren'', and determine whether it needs to \mre more up until the word it is aligned. 
When it decided to \mwr ``gewerkt''; the word ``worked'' would have been scanned in the past; so it should \mwr without an additional \mre.

Next, it is also inevitable that our produced alignments are noisy and do not align all words correctly. It is currently not clear how much this will hurt the performance of the systems in terms of quality and delay due to alignment errors. In the worst case, our oracle will misguide the interpreter to guess target words without appropriate context (lowering quality, lowering delay) or wait for too many words (increasing quality, increasing delay). Both scenarios resulting from this noisy alignment are not catastrophic as it still depends on the interpreter's ability to guess translation outputs without appropriate input context.

\subsection{Transformer and Large Scale Experiments}
\label{subsec:large_scale}

To show that the proposed method also extends to the transformer and larger parallel data, we conduct two experiments. 
The first experiment is changing the LSTM architecture with the transformer architecture similar to \newcite{ma-etal-2019-stacl}.
The second experiment uses 4.5 millions DE to EN parallel sentences from WMT 2015.

\begin{table}[t!]

\begin{subtable}{\textwidth}

\footnotesize

\begin{tabular}{l||ccc|ccc}
& \multicolumn{3}{c}{IWSLT} 
& \multicolumn{3}{c}{WMT} 
\\

& BLEU & AL & AP  & BLEU & AL & AP   \\

\toprule 
wait-$k$ \\ 

\ \ \ $k=1$  
& 11.12 & 2.24 & 0.60
& 13.65 & 1.70 & 0.56
\\

\ \ \ $k=2$ 
& 17.08 & 2.61 & 0.62
& 16.77 & 2.27 & 0.59
\\

\ \ \ $k=3$ 
& 19.52 & 3.33 & 0.66
& 18.30 & 2.95 & 0.62
\\

\ \ \ $k=4$ 
& 21.14 & 4.10 & 0.70
& 19.09 & 3.66 & 0.64
\\

\ \ \ $k=5$ 
& 22.68 & 4.93 & 0.73
& 19.80 & 4.42 & 0.67
\\

\ \ \ $k=6$ 
& 24.16 & 5.68 & 0.76
& 20.70 & 5.28 & 0.70
\\

\ \ \ $k=7$ 
& 26.60 & 6.64 & 0.79
& 21.08 & 6.16 & 0.72
\\

\ \ \ $k=\infty$ 
& 29.53 & 22.46 & 1.00
& 22.51 & 29.49 & 1.00
\\
\midrule

\npi-\simt 
& 22.06 & 1.73 & 0.56
& 17.57 & 3.25 & 0.59
\\

\ \ \ $+ \va'$ 
& 21.40 & 1.69 & 0.56
& 18.18 & 3.17 & 0.59
\\

\ \ \ $+ \va''$ 
& 23.28 & 1.85 & 0.57
& 18.70 & 3.33 & 0.60
\\

\ \ \ $+ \va', \va''$ 
& 23.82 & 1.83 & 0.57
& 18.78 & 3.54 & 0.60
\\

\ \ \ $+ \va', \va'', \vy'$ 
& 24.54 & 1.78 & 0.57
& 19.07 & 3.27 & 0.59
\\

\midrule
Oracle-at-test 
& 30.68 & 1.79 & 0.57
& 27.09 & 2.75 & 0.58 
\\
\bottomrule

\end{tabular}
\end{subtable}
\caption{Results using transformer on IWSLT and WMT corpora.}
\label{tab:trans}
\end{table}

The interpreter is a standard 6 layers encoder-decoder NMT transformer similar to \newcite{NIPS2017_7181}. The programmer consists of single 6 layers encoder transformer with a binary classifier. 
For both networks, we allow attention to attend only to the previous timesteps.
Then we use the program to mask out unseen inputs at each timestep in the interpreter.
Unless otherwise specified, we use the default settings of training transformer as in \newcite{NIPS2017_7181}. 
We employ 30\% dropout to all transformers and 10\% to the embeddings, 16k vocabulary size, an average batch size of 4k, 8k steps learning rate warmup, 50 tokens maximum per sentence during training, for a total of 200k steps.
We use a single pass of parallel scheduled sampling \cite{duckworth2019parallel} for the transformer to generate $\va'$ and $\vy'$ and set the $\vy', \va', \va''$ perturbation rate to be 10\%, 15\%, and 25\%.
Training is completed within 20 hours on a single V100 GPU.\footnote{Because of the limitation of our computational resources, we are unable to use multiple GPUs for larger batch size. Using smaller batch size is known to reduce the overall performance of the transformer.}

Table \ref{tab:trans} presents the Transformer results on IWSLT and WMT. First we see that our transformer results are competitive or better than the LSTM on IWSLT dataset (compare with Table \ref{tab:main}). 
These results are similar to the \newcite{ma-etal-2019-stacl} when comparing LSTM and transformer based architectures in the \simt settings.
Second, we see that our coupled scheduled sampling approach is also able to increase BLEU by up to 1.5 points compared to vanilla \npi-\simt approach while also keeping the delay low (3.27 AL).
The higher AL of the \simt model in WMT compared to IWSLT is likely due to the higher AL of the oracle (2.75 vs. 1.79), which we attribute to the nature of the dataset.
Arguably, this crawled corpus is less suitable for \simt in general, because it contains considerably longer parallel sentences; moreover, the text is less reflective of a real simultaneous interpretation setting as it was built by only matching offline and post-edited texts.
We are  able to see similar improvements using coupled scheduled sampling over vanilla \npi-\simt approach, showing the scalability of our approach.

\section{Related Work}


\newcite{sp16,gu-etal-2017-learning} and \newcite{alinejad-etal-2018-prediction} formulate simultaneous \nmt \  as sequential decision making problem where an agent interacts with the environment (i.e. the underlying \nmt \  model) through \mre/\mwr \ actions. They pre-train the \nmt \  system, while the agent's policy is trained using deep-RL. 

\newcite{arivazhagan2020re} highlights the poor performance of finetuned offline translation model when translating prefixes of input, which is the case of \simt. Their approach uses retranslation strategy where every \mre is performed, a new translation is generated from scratch, allowing revising translation on the fly and mitigating error propagation on the decoder that was attributed to the insufficient evidence when generating past output words. Their approach uses a stability metric which takes number of suffixes revisions made to produce latest translation. This approach involves wait-$k$ inference, which limits number of words that can be emitted by the interpreter during one writing and thus limiting number of suffix revisions at the next writing. This wait-$k$ inference is a heuristic that can be replaced by learning from the oracle.

\newcite{ma-etal-2019-stacl,dalvi-etal-2018-incremental} introduced the fixed wait-$k$ policy, which allows the integrated training of the NMT model wrt the fixed policy, as opposed to the adaptive policy of \newcite{gu-etal-2017-learning, arivazhagan-etal-2019-monotonic} jointly trains an adaptive policy and re-trains the underlying \nmt \  system. 
\newcite{arivazhagan-etal-2019-monotonic, DBLP:conf/acl/ZhengZMH19} produces oracle \mre/\mwr \  actions using a pre-trained \nmt \  model, which is then used to train an adaptive agent based on supervised learning, i.e. behavioural cloning in imitation learning.
Compared to our oracle which is produced merely from word alignment, their method requires a full decoding of training corpus, which is computationally expensive.
These works are different from ours in that: (i) they do not use word alignment to produce the oracle actions, and (ii) they do not use of scheduled sampling.


\section{Conclusion}

This paper proposes a simple and effective way to train a simultaneous translation system to produce low delay translations. 
Our central contribution is to determine a sufficient, if not minimum, amount of inputs to translate each target token. 
This is achieved using word-alignment to create an oracle, which is then used as part of a training algorithm based on imitation learning
to learn coupled policies, for a ``programmer'' which decides when to wait for more input producing translation tokens, and an ``interpreter'' which 
generates the translation.
We show the importance of scheduled sampling during learning, which is crucial to combat exposure bias. 
Overall we show improvements in BLEU score over naively trained systems with modest translation delays.

Future work is needed to better understand the effect of various alignment models and symmetrization methods on the generated oracle. Beyond this, other opportunities include applying the model to the real speech input and applying more sophisticated imitation learning techniques that involves the generated trajectories of both ``interpreter'' and ``programmer''.

\section*{Acknowledgment}
We thank anonymous reviewers for their valuable comments that helped to make this paper considerably better. This work is supported by the Australian Research Council (ARC DP160102686) and an Amazon Research Award. 

\bibliographystyle{acl_natbib}
\bibliography{main_sim_nmt}

\begin{thebibliography}{27}
\expandafter\ifx\csname natexlab\endcsname\relax\def\natexlab#1{#1}\fi

\bibitem[{Alinejad et~al.(2018)Alinejad, Siahbani, and
  Sarkar}]{alinejad-etal-2018-prediction}
Ashkan Alinejad, Maryam Siahbani, and Anoop Sarkar. 2018.
\newblock Prediction improves simultaneous neural machine translation.
\newblock In \emph{Proceedings of the Conference on Empirical Methods in
  Natural Language Processing}.

\bibitem[{Arivazhagan et~al.(2019)Arivazhagan, Cherry, Macherey, Chiu, Yavuz,
  Pang, Li, and Raffel}]{arivazhagan-etal-2019-monotonic}
Naveen Arivazhagan, Colin Cherry, Wolfgang Macherey, Chung-Cheng Chiu, Semih
  Yavuz, Ruoming Pang, Wei Li, and Colin Raffel. 2019.
\newblock Monotonic infinite lookback attention for simultaneous machine
  translation.
\newblock In \emph{Proceedings of the 57th Annual Meeting of the Association
  for Computational Linguistics}.

\bibitem[{Arivazhagan et~al.(2020)Arivazhagan, Cherry, Macherey, and
  Foster}]{arivazhagan2020re}
Naveen Arivazhagan, Colin Cherry, Wolfgang Macherey, and George Foster. 2020.
\newblock Re-translation versus streaming for simultaneous translation.
\newblock \emph{arXiv preprint arXiv:2004.03643}.

\bibitem[{Bahdanau et~al.(2015)Bahdanau, Cho, and
  Bengio}]{DBLP:journals/corr/BahdanauCB14}
Dzmitry Bahdanau, Kyunghyun Cho, and Yoshua Bengio. 2015.
\newblock \href {http://arxiv.org/abs/1409.0473} {Neural machine translation by
  jointly learning to align and translate}.
\newblock In \emph{3rd International Conference on Learning Representations,
  {ICLR} 2015, San Diego, CA, USA, May 7-9, 2015, Conference Track
  Proceedings}.

\bibitem[{Bengio et~al.(2015)Bengio, Vinyals, Jaitly, and
  Shazeer}]{NIPS2015_5956}
Samy Bengio, Oriol Vinyals, Navdeep Jaitly, and Noam Shazeer. 2015.
\newblock Scheduled sampling for sequence prediction with recurrent neural
  networks.
\newblock In C.~Cortes, N.~D. Lawrence, D.~D. Lee, M.~Sugiyama, and R.~Garnett,
  editors, \emph{Advances in Neural Information Processing Systems 28}, pages
  1171--1179. Curran Associates, Inc.

\bibitem[{Brown et~al.(1993)Brown, Pietra, Pietra, and Mercer}]{Brown1994}
Peter~F. Brown, Stephen~Della Pietra, Vincent J.~Della Pietra, and Robert~L.
  Mercer. 1993.
\newblock The mathematics of statistical machine translation: Parameter
  estimation.
\newblock \emph{Computational Linguistics}, 19(2):263--311.

\bibitem[{Cettolo et~al.(2012)Cettolo, Girardi, and
  Federico}]{cettoloEtAl:EAMT2012}
Mauro Cettolo, Christian Girardi, and Marcello Federico. 2012.
\newblock Wit$^3$: Web inventory of transcribed and translated talks.
\newblock In \emph{Proceedings of the 16$^{th}$ Conference of the European
  Association for Machine Translation (EAMT)}, pages 261--268, Trento, Italy.

\bibitem[{Dalvi et~al.(2018)Dalvi, Durrani, Sajjad, and
  Vogel}]{dalvi-etal-2018-incremental}
Fahim Dalvi, Nadir Durrani, Hassan Sajjad, and Stephan Vogel. 2018.
\newblock Incremental decoding and training methods for simultaneous
  translation in neural machine translation.
\newblock In \emph{Meeting of the North American Chapter of the Association for
  Computational Linguistics (NAACL)}, pages 493--499.

\bibitem[{Duckworth et~al.(2019)Duckworth, Neelakantan, Goodrich, Kaiser, and
  Bengio}]{duckworth2019parallel}
Daniel Duckworth, Arvind Neelakantan, Ben Goodrich, Lukasz Kaiser, and Samy
  Bengio. 2019.
\newblock Parallel scheduled sampling.
\newblock \emph{arXiv preprint arXiv:1906.04331}.

\bibitem[{Dyer et~al.(2013)Dyer, Chahuneau, and Smith}]{dyer-etal-2013-simple}
Chris Dyer, Victor Chahuneau, and Noah~A. Smith. 2013.
\newblock A simple, fast, and effective reparameterization of {IBM} model 2.
\newblock In \emph{Proceedings of the 2013 Conference of the North {A}merican
  Chapter of the Association for Computational Linguistics: Human Language
  Technologies}, pages 644--648.

\bibitem[{Grissom~II et~al.(2014)Grissom~II, He, Boyd-Graber, Morgan, and
  Daum{\'e}~III}]{grissom-ii-etal-2014-dont}
Alvin Grissom~II, He~He, Jordan Boyd-Graber, John Morgan, and Hal
  Daum{\'e}~III. 2014.
\newblock Don{'}t until the final verb wait: Reinforcement learning for
  simultaneous machine translation.
\newblock In \emph{Proceedings of the 2014 Conference on Empirical Methods in
  Natural Language Processing ({EMNLP})}, pages 1342--1352.

\bibitem[{Gu et~al.(2017)Gu, Neubig, Cho, and Li}]{gu-etal-2017-learning}
Jiatao Gu, Graham Neubig, Kyunghyun Cho, and Victor~O.K. Li. 2017.
\newblock \href {https://www.aclweb.org/anthology/E17-1099} {Learning to
  translate in real-time with neural machine translation}.
\newblock In \emph{Proceedings of the 15th Conference of the {E}uropean Chapter
  of the Association for Computational Linguistics: Volume 1, Long Papers},
  pages 1053--1062, Valencia, Spain. Association for Computational Linguistics.

\bibitem[{Kingma and Ba(2015)}]{DBLP:journals/corr/KingmaB14}
Diederik~P. Kingma and Jimmy Ba. 2015.
\newblock \href {http://arxiv.org/abs/1412.6980} {Adam: {A} method for
  stochastic optimization}.
\newblock In \emph{3rd International Conference on Learning Representations,
  {ICLR} 2015, San Diego, CA, USA, May 7-9, 2015, Conference Track
  Proceedings}.

\bibitem[{Koehn et~al.(2003)Koehn, Och, and
  Marcu}]{koehn-etal-2003-statistical}
Philipp Koehn, Franz~J. Och, and Daniel Marcu. 2003.
\newblock \href {https://www.aclweb.org/anthology/N03-1017} {Statistical
  phrase-based translation}.
\newblock In \emph{Proceedings of the 2003 Human Language Technology Conference
  of the North {A}merican Chapter of the Association for Computational
  Linguistics}, pages 127--133.

\bibitem[{Kudo and Richardson(2018)}]{kudo2018sentencepiece}
Taku Kudo and John Richardson. 2018.
\newblock Sentencepiece: A simple and language independent subword tokenizer
  and detokenizer for neural text processing.
\newblock \emph{arXiv preprint arXiv:1808.06226}.

\bibitem[{Luong et~al.(2015)Luong, Pham, and
  Manning}]{luong-etal-2015-effective}
Thang Luong, Hieu Pham, and Christopher~D. Manning. 2015.
\newblock \href {https://doi.org/10.18653/v1/D15-1166} {Effective approaches to
  attention-based neural machine translation}.
\newblock In \emph{Proceedings of the 2015 Conference on Empirical Methods in
  Natural Language Processing}, pages 1412--1421, Lisbon, Portugal. Association
  for Computational Linguistics.

\bibitem[{Ma et~al.(2019)Ma, Huang, Xiong, Zheng, Liu, Zheng, Zhang, He, Liu,
  Li, Wu, and Wang}]{ma-etal-2019-stacl}
Mingbo Ma, Liang Huang, Hao Xiong, Renjie Zheng, Kaibo Liu, Baigong Zheng,
  Chuanqiang Zhang, Zhongjun He, Hairong Liu, Xing Li, Hua Wu, and Haifeng
  Wang. 2019.
\newblock \href {https://doi.org/10.18653/v1/P19-1289} {{STACL}: Simultaneous
  translation with implicit anticipation and controllable latency using
  prefix-to-prefix framework}.
\newblock In \emph{Proceedings of the 57th Annual Meeting of the Association
  for Computational Linguistics}, pages 3025--3036, Florence, Italy.
  Association for Computational Linguistics.

\bibitem[{Murray and Chiang(2018)}]{murray-chiang-2018-correcting}
Kenton Murray and David Chiang. 2018.
\newblock \href {https://doi.org/10.18653/v1/W18-6322} {Correcting length bias
  in neural machine translation}.
\newblock In \emph{Proceedings of the Third Conference on Machine Translation:
  Research Papers}, pages 212--223, Brussels, Belgium. Association for
  Computational Linguistics.

\bibitem[{Papineni et~al.(2002)Papineni, Roukos, Ward, and
  Zhu}]{papineni-etal-2002-bleu}
Kishore Papineni, Salim Roukos, Todd Ward, and Wei-Jing Zhu. 2002.
\newblock \href {https://doi.org/10.3115/1073083.1073135} {{B}leu: a method for
  automatic evaluation of machine translation}.
\newblock In \emph{Proceedings of the 40th Annual Meeting of the Association
  for Computational Linguistics}, pages 311--318, Philadelphia, Pennsylvania,
  USA. Association for Computational Linguistics.

\bibitem[{Post(2018)}]{post2018call}
Matt Post. 2018.
\newblock \href {https://doi.org/10.18653/v1/W18-6319} {A call for clarity in
  reporting {BLEU} scores}.
\newblock In \emph{Proceedings of the Third Conference on Machine Translation:
  Research Papers}, pages 186--191, Brussels, Belgium. Association for
  Computational Linguistics.

\bibitem[{Ross et~al.(2011)Ross, Gordon, and Bagnell}]{pmlr-v15-ross11a}
Stephane Ross, Geoffrey Gordon, and Drew Bagnell. 2011.
\newblock A reduction of imitation learning and structured prediction to
  no-regret online learning.
\newblock In \emph{Proceedings of the Fourteenth International Conference on
  Artificial Intelligence and Statistics}, pages 627--635.

\bibitem[{Satija and Pineau(2016)}]{sp16}
Harsh Satija and Joelle Pineau. 2016.
\newblock Simultaneous machine translation using deep reinforcement learning.
\newblock In \emph{Proceedings of the Abstraction in Reinforcement Learning
  Workshop}.

\bibitem[{Torabi et~al.(2019)Torabi, Warnell, and Stone}]{ijcai2019-882}
Faraz Torabi, Garrett Warnell, and Peter Stone. 2019.
\newblock Recent advances in imitation learning from observation.
\newblock In \emph{Proceedings of the Twenty-Eighth International Joint
  Conference on Artificial Intelligence, {IJCAI-19}}, pages 6325--6331.
  International Joint Conferences on Artificial Intelligence Organization.

\bibitem[{Tyers and Alperen(2010)}]{tyers2010south}
Francis~M Tyers and Murat~Serdar Alperen. 2010.
\newblock South-east european times: A parallel corpus of balkan languages.
\newblock In \emph{Proceedings of the LREC Workshop on Exploitation of
  Multilingual Resources and Tools for Central and (South-) Eastern European
  Languages}, pages 49--53.

\bibitem[{Vaswani et~al.(2017)Vaswani, Shazeer, Parmar, Uszkoreit, Jones,
  Gomez, Kaiser, and Polosukhin}]{NIPS2017_7181}
Ashish Vaswani, Noam Shazeer, Niki Parmar, Jakob Uszkoreit, Llion Jones,
  Aidan~N Gomez, \L~ukasz Kaiser, and Illia Polosukhin. 2017.
\newblock \href
  {http://papers.nips.cc/paper/7181-attention-is-all-you-need.pdf} {Attention
  is all you need}.
\newblock In I.~Guyon, U.~V. Luxburg, S.~Bengio, H.~Wallach, R.~Fergus,
  S.~Vishwanathan, and R.~Garnett, editors, \emph{Advances in Neural
  Information Processing Systems 30}, pages 5998--6008. Curran Associates, Inc.

\bibitem[{Zheng et~al.(2019{\natexlab{a}})Zheng, Zheng, Ma, and
  Huang}]{DBLP:journals/corr/abs-1909-01559}
Baigong Zheng, Renjie Zheng, Mingbo Ma, and Liang Huang. 2019{\natexlab{a}}.
\newblock Simpler and faster learning of adaptive policies for simultaneous
  translation.
\newblock In \emph{Proceedings of the 2019 Conference on Empirical Methods in
  Natural Language Processing and the 9th International Joint Conference on
  Natural Language Processing}, pages 1349--1354.

\bibitem[{Zheng et~al.(2019{\natexlab{b}})Zheng, Zheng, Ma, and
  Huang}]{DBLP:conf/acl/ZhengZMH19}
Baigong Zheng, Renjie Zheng, Mingbo Ma, and Liang Huang. 2019{\natexlab{b}}.
\newblock Simultaneous translation with flexible policy via restricted
  imitation learning.
\newblock In \emph{Proceedings of the 57th Conference of the Association for
  Computational Linguistics, {ACL} 2019, Florence, Italy, July 28- August 2,
  2019, Volume 1: Long Papers}, pages 5816--5822.

\end{thebibliography}

\end{document}